\documentclass{article}

\usepackage{microtype}
\usepackage{graphicx}
\usepackage{subfigure}
\usepackage{booktabs} 
\usepackage{supertabular, booktabs} 
\usepackage{array}

\usepackage{hyperref}

\usepackage[accepted]{icml2018}

\icmltitlerunning{Fashion-Gen: The Generative Fashion Dataset and Challenge}

\begin{document}

 \newcolumntype{L}[1]{>{\raggedright\arraybackslash}p{#1}}
 \newcolumntype{C}[1]{>{\centering\arraybackslash}p{#1}}
 \newcolumntype{R}[1]{>{\raggedleft\arraybackslash}p{#1}}

\twocolumn[
\icmltitle{Fashion-Gen: The Generative Fashion Dataset and Challenge}



\icmlsetsymbol{equal}{*}

\begin{icmlauthorlist}
\icmlauthor{Negar Rostamzadeh}{element}
\icmlauthor{Seyedarian Hosseini}{element,mo}
\icmlauthor{Thomas Boquet}{element}
\icmlauthor{Wojciech Stokowiec}{element}
\icmlauthor{Ying Zhang}{element}
\icmlauthor{Christian Jauvin}{element}
\icmlauthor{Chris Pal}{ed,element}
\end{icmlauthorlist}

\icmlaffiliation{element}{Element AI}
\icmlaffiliation{mo}{MILA-University of Montreal}
\icmlaffiliation{ed}{Ecole Polytechnique of Montreal}

\icmlcorrespondingauthor{Negar Rostamzadeh}{negar@elementai.com}

\icmlkeywords{Machine Learning, ICML}

\vskip 0.3in
]



\printAffiliationsAndNotice{}  

\begin{abstract}
We introduce a new dataset of 293,008 high definition (1360 x 1360 pixels) fashion images paired with item descriptions provided by professional stylists. Each item is photographed from a variety of angles. We provide baseline results on 1) high-resolution image generation, and 2) image generation conditioned on the given text descriptions. We invite the community to improve upon these baselines. In this paper we also outline the details of a challenge that we are launching based upon this dataset.
\end{abstract}

\section{Introduction}
Machine learning has recently been employed in many applications pertaining to the fashion industry. The use cases range from style matching~\cite{bossard2012apparel, kalantidis2013getting, liu2016deepfashion}, recommendation systems in e-commerce sites~\cite{chen2012describing, xiao2015learning, kiapour2015buy, chen2015deep, simo2015neuroaesthetics}, trend prediction, the ability for customers to virtually try on clothes~\cite{han2017viton}, and clothing type classification~\cite{liu2012street, liang2016clothes, veit2015learning, zhu2017your}.

The availability of large-scale datasets such as DeepFashion~\cite{liu2016deepfashion}  
has fueled recent progress in applying deep learning to fashion tasks. However, there are still many aspects of the industry that computer vision methods have not been applied to. In this paper we explore the task of assisting fashion designers to share their ideas with others by translating verbal descriptions to images. Thus, given a description of a particular item, we generate images of clothes and accessories matching the description. 

\begin{figure}
\centering
\begin{subfigure}
  \centering
  \includegraphics[width=7.2cm]{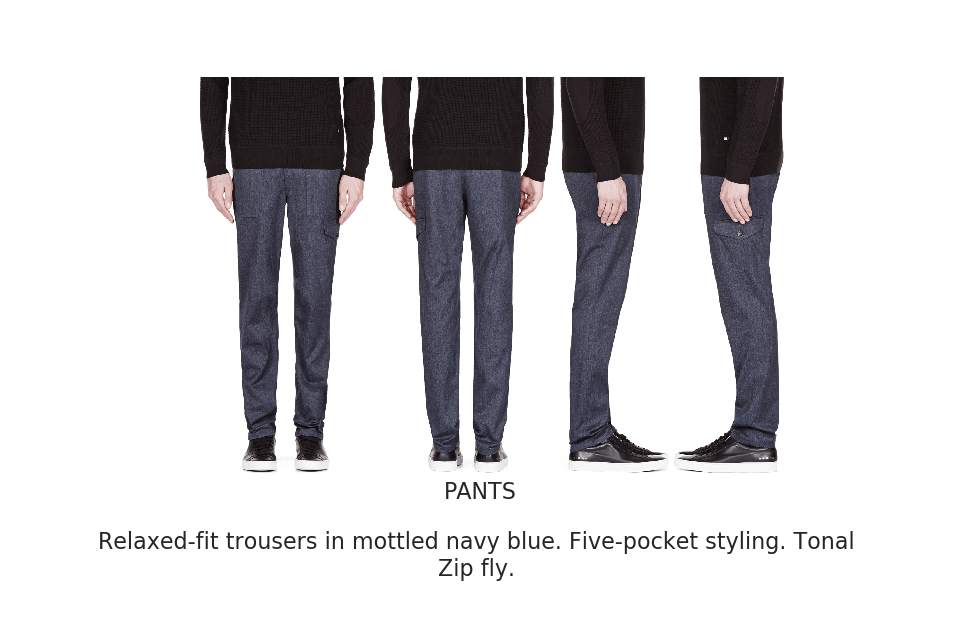}
\end{subfigure}
 \\
 (a)
 \\
\begin{subfigure}
\centering
\includegraphics[width=7.2cm]{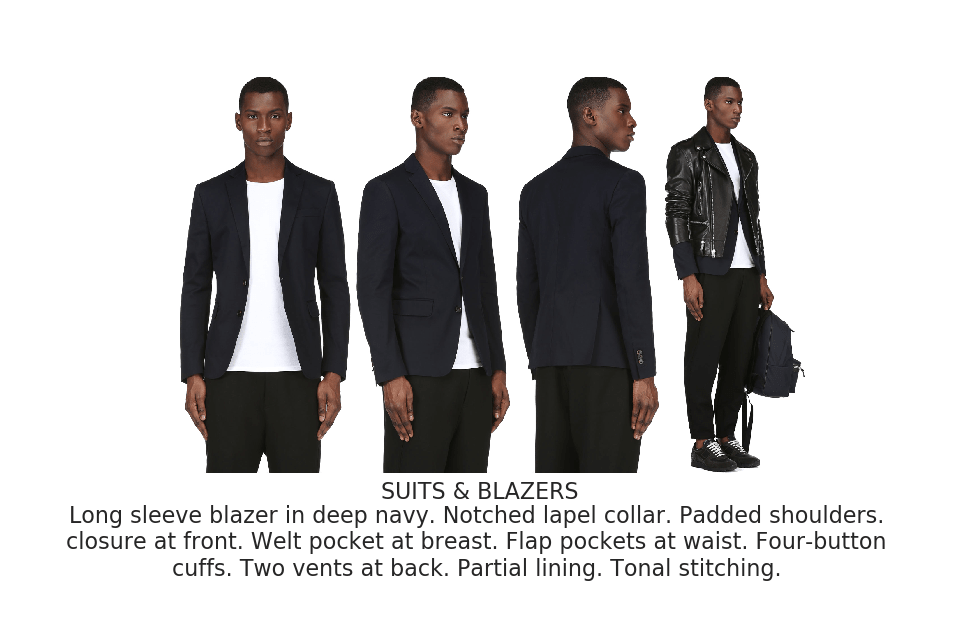}
\end{subfigure}
 \\
 (b)
 \\
\begin{subfigure}
\centering
\includegraphics[width=7.2cm]{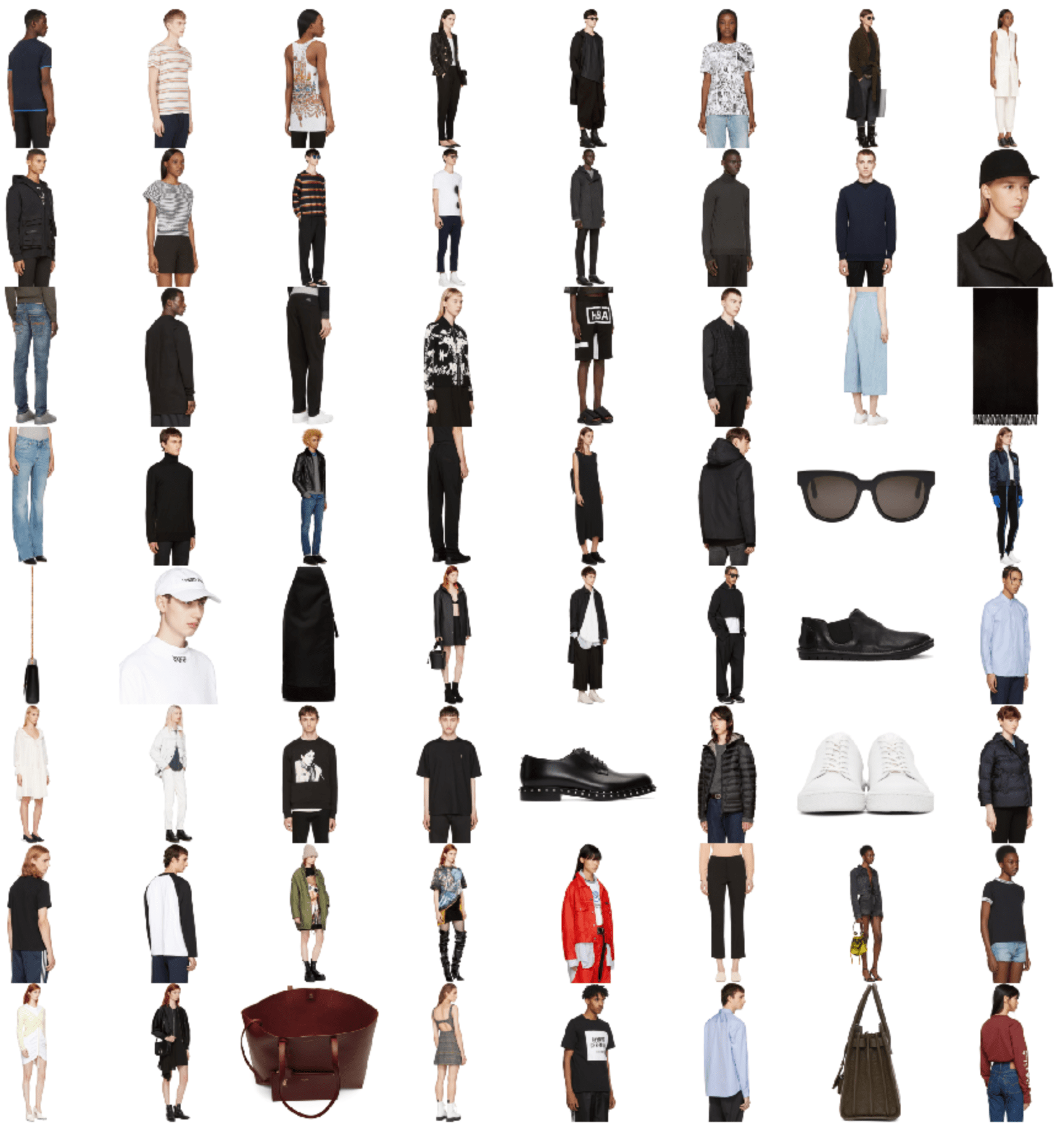}
\end{subfigure}
 \\
 (c)
 \\
\caption{Pictures a, b and c present samples of the dataset. Each description is associated with all the images below it. And each item \textit{ie. a, b} is photographed from different angles. We also provide each image's attributes, and its relationship to other objects in the dataset}
\label{dataset_Sample1} 
\end{figure}

To explore these research directions we introduce here a new dataset of almost 300k high definition training images of clothes, and accessories accompanied by detailed design descriptions.
%
%
%
Each description is provided by professional designers and contains fine-grained design details. Each product is photographed from multiple angles against a standardized background under consistent lighting conditions and annotated with matching items recommended by a stylist. See Figure~\ref{dataset_Sample1}  for examples.

In this paper we provide: 1) statistical details of the dataset, 2) detailed comparisons with existing datasets, 3) an introduction to the competition that we are launching on the task of text to image generation, with a brief explanation of the competition criteria and evaluation process, and 4) high-resolution image generation results using an approach based on the progressive growing of GANs \cite{karras2017progressive}, and text-to-image translation results using StackGAN-v1 \cite{zhang2017stackgan}, and StackGAN-v2 \cite{huang2017stacked}. 

The paper is organized as follows: Section~\ref{related} discusses related work. Section~\ref{dataset} introduces the Fashion dataset, describes the collection procedure, and provides a statistical analysis of the dataset with details of our newly introduced challenge.\footnote{The competition is part of the first workshop of Computer Vision for Fashion, Art and Design at ECCV. The challenge website is \url{https://fashion-gen.com/}} 
In Section~\ref{experiments}, we describe baseline approaches and the evaluation process, including human evaluation. Section~\ref{conclusion}, concludes the paper and discusses future work.

\section{Related Work}\label{related}
We first provide a summary of generative models used in text-to-image synthesis and then discuss related datasets.
\subsection{Applications of Generative models}
Generative Adversarial Networks~\cite{goodfellow2014generative} have been used in a wide range of applications, including
photo-realistic image super-resolution~\cite{ledig2016photo, sonderby2016amortised}, video generation~\cite{denton2018stochastic, denton2017unsupervised}, inpainting~\cite{belghazi2018hierarchical}, image-to-image translation~\cite{isola2017image, zhu2017unpaired,taigman2016unsupervised} and text-to-image synthesis~\cite{zhang2017stackgan, huang2017stacked, reed2016generative, reed2016learning, zhang2018photographic}. 

Although state of the art  generative models can already generate polished realistic images~\cite{karras2017progressive}, conditional generation and translation tasks are still far from high quality. We hypothesize that this shortcoming is due to a shortage of large, clean datasets.

\subsection{Related datasets}
To the best of our knowledge none of the currently used datasets for text-to-image synthesis were  collected specifically for the purposes of exploring the text-to-image synthesis task.
Below we discuss existing datasets that have been used for text-to-image and attributes-to-image synthesis and focus on a specific set of attributes which are important for image synthesis.

\begin{table*}[]
\tiny
\centering
\caption{Comparison of datasets}
\label{my-label}
  \begin{tabular}{@{}lcccccccl@{}}
   & \multicolumn{1}{l}{Number of images} & \multicolumn{1}{c}{Resolution}                  & \multicolumn{1}{l}{Description} & \multicolumn{1}{l}{Binary attributes} & \multicolumn{1}{l}{Categories} & \multicolumn{1}{l}{Poses} & \multicolumn{1}{l}{Number of items} &  \\
  \hline
  CelebA                                & 202,599                              & 43x55  to 6732x8984                            & no                              & 40                                    & no                             & multiple                  & 10,177                                        \\
  CelebA-HQ                             & 30,000                               & 1024x1024                                       & no                              & 40                                    & no                             & multiple                  & unknown                                  \\
  DeepFashion - Fashion Image Synthesis & 78,979                               & 300x300                                             & multiple                        & 1000                                  & 50                             & multiple                  & unknown                                \\
  MS COCO                               & 328,000                              & varying sizes                                   & 5 per image                     & no                                    & 80                             & single                    & unknown                                  \\
  Caltech-UCSD Birds-200-2011           & 11,788                               & varying sizes & no                              & 312                                   & 200                            & single                    & unknown                                       \\
  Flowers Oxford-102                    & 8189                                 & varying sizes & no                              & no                                    & 102                            & single                    & unknown                             &   \\
  \textbf{Fashion dataset (ours)}                 & 325,536                              & 1360x1360                                       & yes                             & no                                    & 48                             & multiple                  & 78850                                    \\   
\bottomrule
\end{tabular}

\end{table*}

\begin{figure}
    \centering
	\includegraphics[height=8cm]{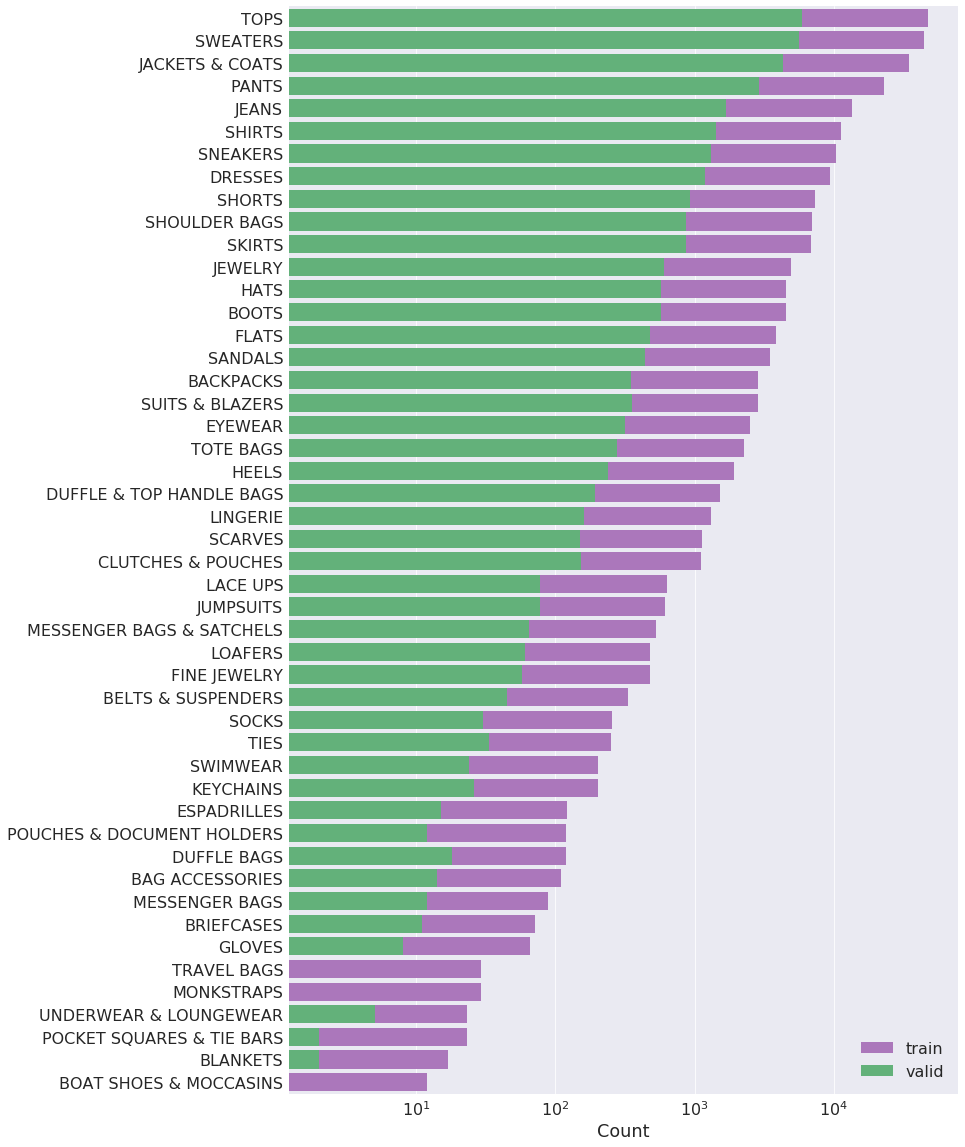}
	\caption{Distribution of the data per category. Note that the $x$ axis is in log scale.}
    	\label{categories}
\end{figure}

\textbf{Caltech-UCSD Birds-200-2011}~\cite{reed2016generative} was originally created for categorizing bird species, localizing their body-parts and classifying attributes. The dataset consists of 12k images , depicting 200 bird species with 28 attributes. More recent work employs this dataset for image synthesis tasks conditioned on text describing the attributes.
%
%
\\
\textbf{MS COCO}~\cite{lin2014microsoft} was originally created as a benchmark for image captioning. While some works use this dataset for text-to-image generation tasks, the generated images miss fine-grained details and only capture high-level information. This is due to the fact that the textual descriptions are very high-level.
%
%
%
%
\\
\textbf{Flowers Oxford-102}~\cite{nilsback2008automated} consists of 102 categories of flowers and was proposed for the task of fine-grained image classification. \citep{reed2016generative} collected 5 descriptions for each image in the dataset to augment it for the task of text to image generation. 
%
%
\\
\textbf{CelebA}~\cite{liu2015deep} contains pictures of 10k celebrities, with 20 images per person (200k images in total). Each image in CelebA is annotated with 40 attributes.
\\
\textbf{DeepFashion}~\cite{liu2016deepfashion} contains over 200k images downloaded from a variety of sources, with varying image sizes, qualities and poses. Each image is annotated with a range of attributes. This publicly available dataset was mainly employed for the task of cloth retrieval and classification. As an extension of the dataset on the task of text-to-image generation, 79k images from the dataset were later annotated with more descriptive text~\cite{zhu2017your}.

\section{Our Fashion Dataset}\label{dataset}
\begin{figure}
    \centering
    \includegraphics[height=8cm]{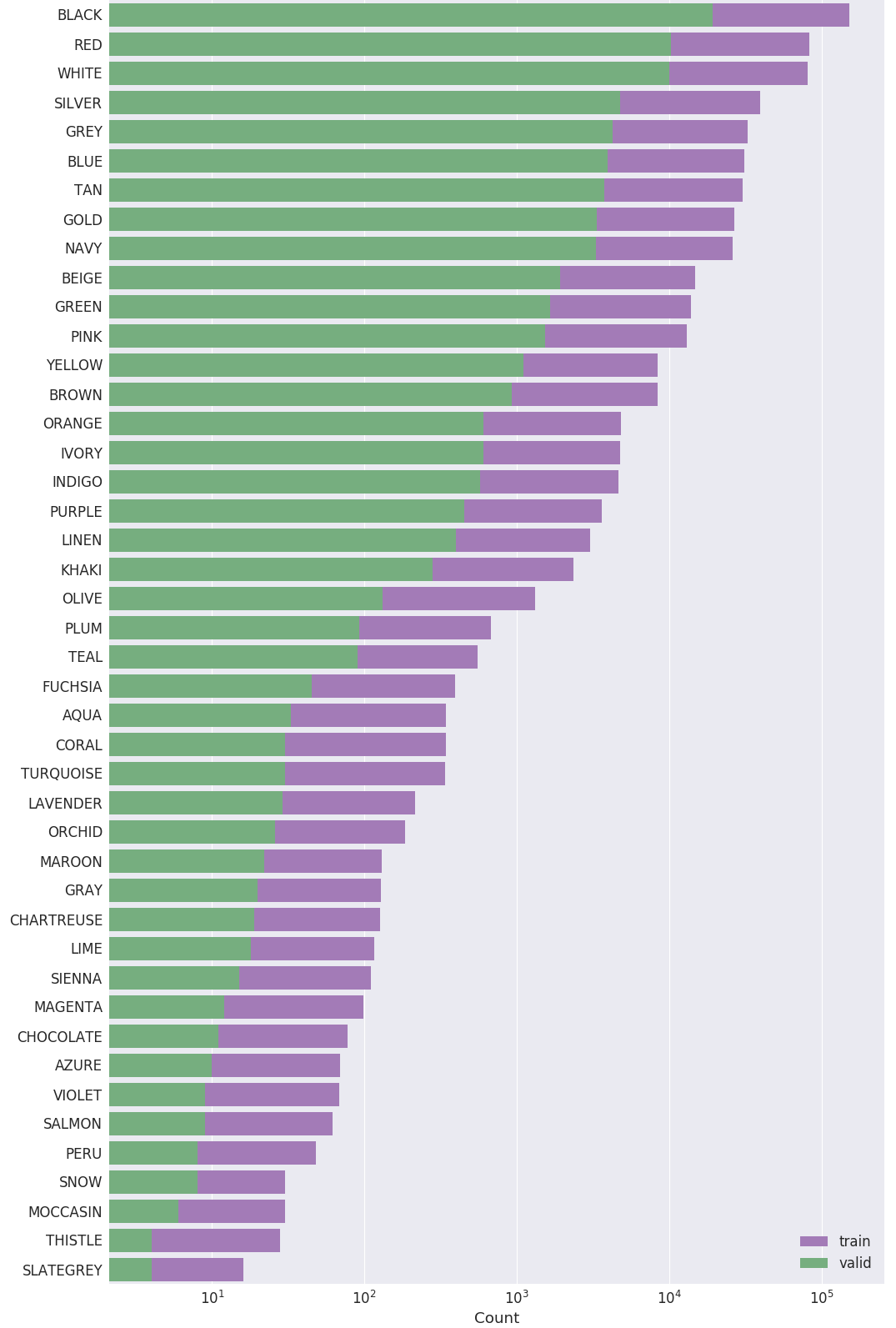}
    \caption{Distribution of the data based on colors. Note that the $x$ axis is in log scale.}
    \label{color}
\end{figure}

\begin{figure*}
    \centering
    \includegraphics[width=0.8\textwidth]{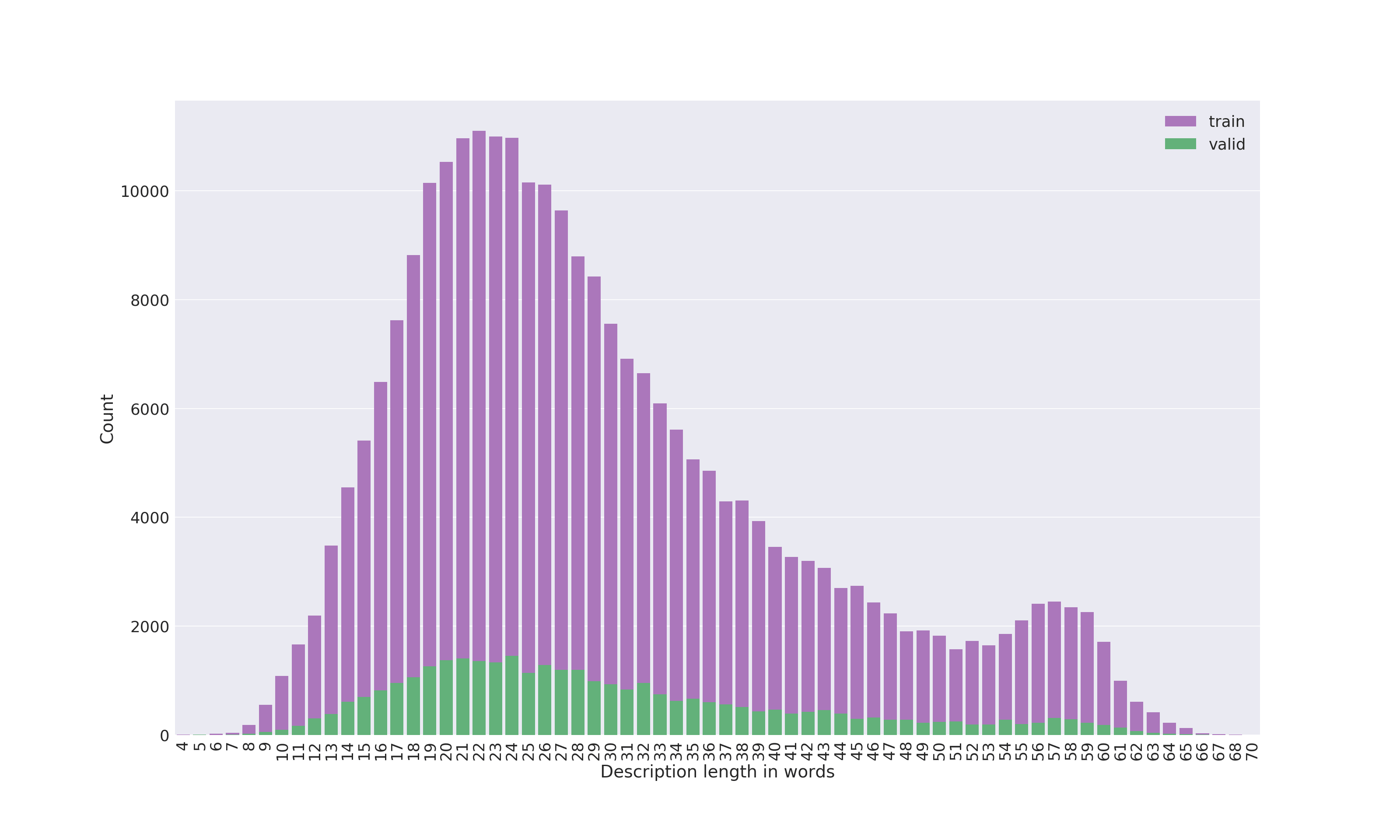}
    \caption{Distribution of the description lengths.}
    \label{desc_length}
\end{figure*}

The advantages of our new Fashion dataset over other contemporary datasets are as follows:

\begin{itemize}
\item The dataset consists of $293,008$ images ($260,480$ images for training, $32,528$ for validation, $32,528$ for test), which is larger than other available datasets for the task of text to image translation.
%
%
\item We provide full HD images photographed under consistent studio conditions. There are no other datasets with comparable resolution and consistent photographing condition.
\item All fashion items are photographed from $1$ to $6$ different angles depending on the category of the item. To our knowledge, this is the first dataset of this scale consisting of multiple angles of each item.
%
%
\item Each product belongs to a main category and a more fine-grained category (\textit{i.e: subcategory}). There are $48$ main categories, and $121$ fine-grained categories in the dataset. The name and density of each category is plotted in~\ref{categories}. Table~\ref{subcat} presents the number of images by category and subcategory.

\item Each fashion item is paired with paragraph-length descriptive captions sourced from experts (\textit{professional designers}). The distribution of the length of descriptions is presented in Figure~\ref{desc_length}.

\item For each item, we also provide metadata such as stylist-recommended matched items, the fashion season, designer and the brand. We also provide the distribution of colors extracted from the text description presented in Figure~\ref{color}
%
%
%
%
\end{itemize}

\section{Our Challenge}\label{challenge}
In addition to releasing a rich dataset, we are launching a challenge that uses our Fashion dataset for the task of text-to-image synthesis. 
To the best of our knowledge this is the first challenge on this task. Additionally, we encourage participants to take advantage of all information in the dataset, e.g. such as pose or category. 
We provide a framework that enables researchers to easily compare the performance of their models with an evaluation metric based on an \textit{Inception Score}~\cite{salimans2016improved}. The inception model we use for the experiments we present in Section \ref{experiments} was trained on the training set for classifying the images into the categories presented in Figure~\ref{categories}. For the final challenge evaluation we will also provide inception scores from a model trained on the test set. However, there are a number of issues to consider when using inception scores for evaluating generative models \cite{barratt2018note}. For example, different implementations of the same model trained on the same dataset can result in significant differences in Inception scores. For these and other reasons our challenge will also provide a human evaluation as we outline below. 
%
%

Our automated evaluation platform for the challenge computes and displays the Inception score for each submission and compiles the best scores in a leader-board.
We provide a comprehensive template and an easy to use service to submit a docker container that runs code, and evaluate the performance on an Amazon Web Services cloud instance. Our test set, which won't be released, consists of descriptions of clothing items and is integrated at runtime in the challengers' docker container. 
%
%
%
%
\\
\\
\textbf{Human Evaluation setup}:
\\
Inception scores do not consider the correlation between text and the given image. As such, the competition results will also be evaluated by humans. Since inception scores also have other issues \cite{barratt2018note} as discussed above, the competition winner will be determined based on this human evaluation. During the human evaluation phase, a fixed subset of the test-set 
will be randomly selected and the
corresponding images will be given to a human evaluation system.
Each human-evaluator will be given a text and $5$ images generated by each submission. The person's task will be to rank these sets of images into the first, second, and third best set with respect to the given text. Each task of this nature will be given to $10$ different human-evaluators. The scores given to each image set will then be 
aggregated to compute final scores under the human evaluation. 

\section{Experiments with the Dataset}\label{experiments}
In this section, we present two sets of experiments: 1) Generating high-resolution images by using the progressive GAN (P-GAN) growing technique of ~\citet{karras2017progressive}, and 2) text-to-image synthesis using StackGAN-v1 ~\cite{zhang2017stackgan} and StackGAN-v2 ~\cite{DBLP:journals/corr/abs-1710-10916}.
\\
\subsection{Generating high-resolution images using P-GANs}
The primary idea of Progressive Growing of GANs~\cite{karras2017progressive} is to grow the generator and discriminator gradually and in a symmetric manner in order to produce high-resolution images. P-GAN starts with very low-resolution images and each new layer of the model improves quality and adds fine-grained details to the image generated in the prior stage. Experiments on the CelebA dataset~\cite{liu2015deep} showed promising results and we similarly employ P-GANs to generate $1024\times 1024$ images using our fashion dataset as training data.
To do this, we follow the same experimental setup and architectural details of the original P-GAN paper~\cite{karras2017progressive}\footnote{Using code provided by the authors of the P-GAN paper~\cite{karras2017progressive}: \url{https://github.com/tkarras/progressive_growing_of_gans}} 

\begin{figure*}
\centering
\includegraphics[width=\textwidth]{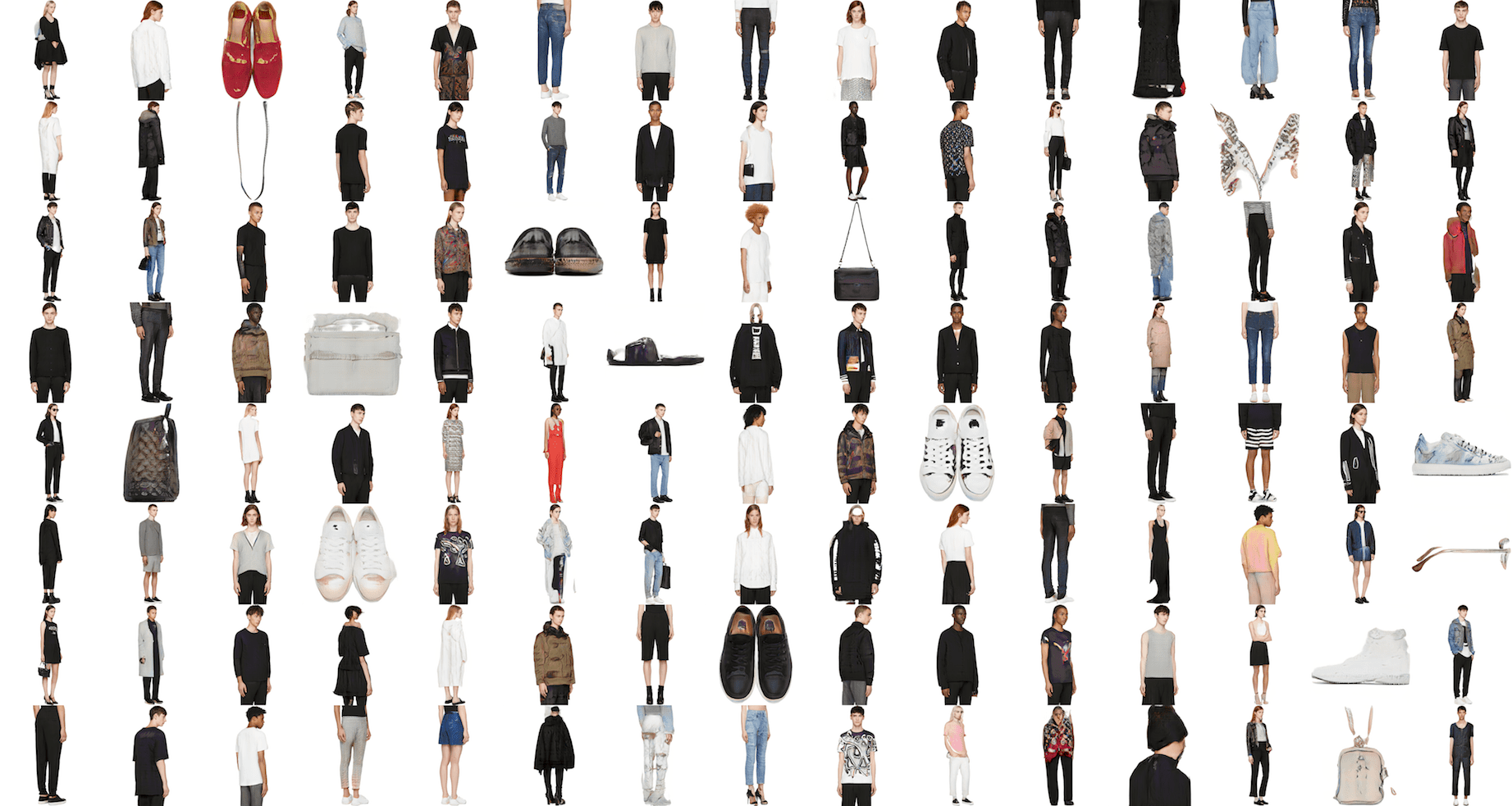}
\caption{Generated images from the P-GAN approach~\cite{karras2017progressive}}
\label{pgan_lr}
\end{figure*}

Figure~\ref{pgan_lr} shows examples of images generated by P-GAN. The images exhibit global coherence and span a variety of poses and attributes ranging from color and category to accessory textures and characteristics of fashion designs.

\begin{figure}[h]
    \centering
    \includegraphics[height=10cm]{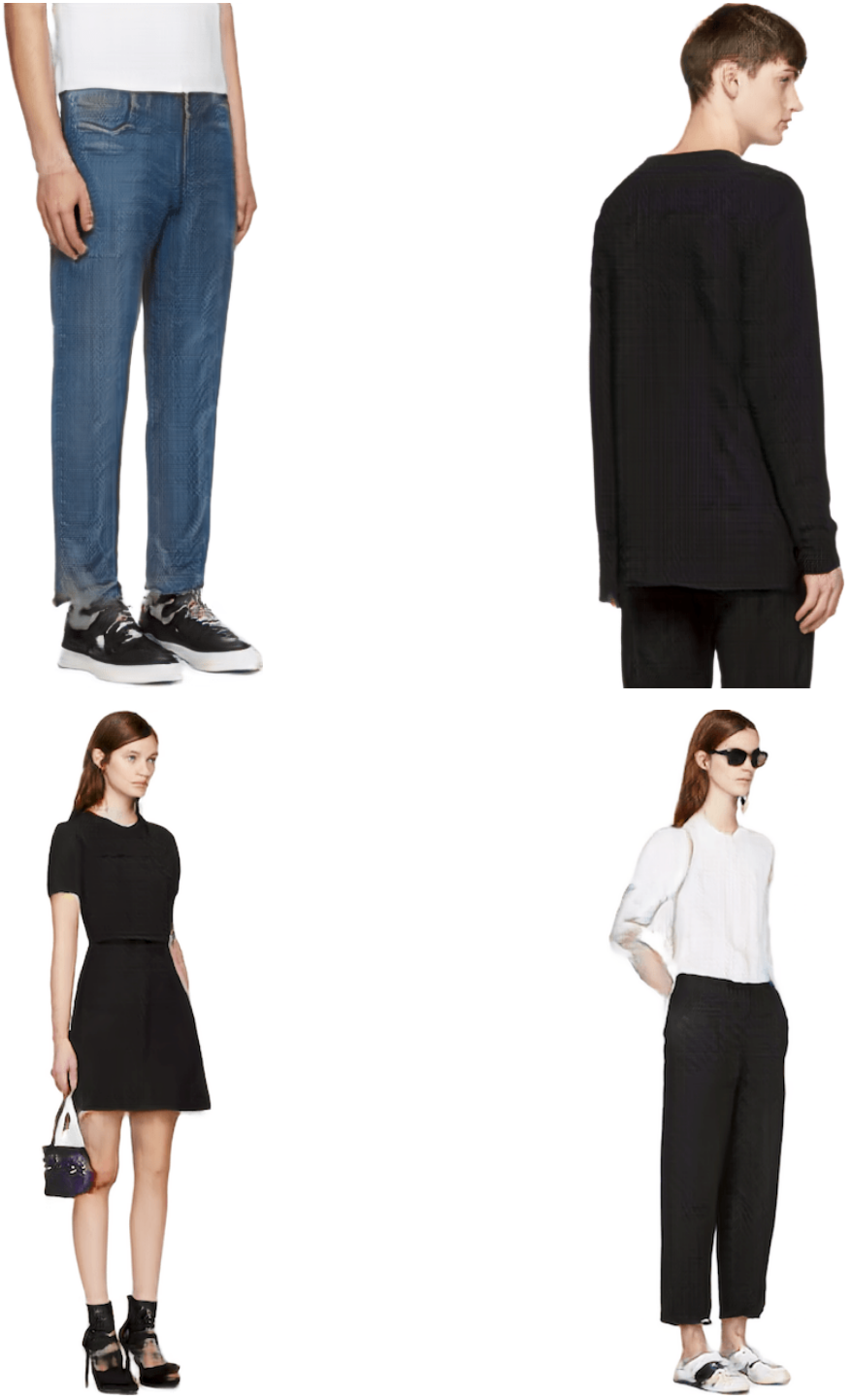}
    \caption{Images generated by the P-GAN approach~\cite{karras2017progressive}}
    \label{pgan}
\end{figure}

In order to quantitatively evaluate the quality of our generated images, we compute the Inception score for the down-sampled version ($256\times 256$) of our generated images (See Table~\ref{inception}). The Inception score of the generated images using P-GANs is very close to the that of the original images, presented in Figure~\ref{pgan}.

\subsection{Text-to-Image synthesis:}
We employed two architectures: \textbf{StackGAN-v1}~\citep{zhang2017stackgan} and \textbf{StackGAN-v2}~\citep{DBLP:journals/corr/abs-1710-10916} to generate images conditioned on the their description. 

\textbf{StackGAN-v1} decomposes conditional image generation into two stages. First, the \textit{Stage-I} GAN sketches a low resolution image ($64\times 64$) with the overall shape and colors of the image conditioned on the text and a random noise vector. Subsequently, the \textit{Stage-II} GAN refines this low-resolution image conditioned on the results of the first stage and the same text embeddings, and generates a $256\times 256$ image.

\textbf{StackGAN-v2} follows a similar architecture consisting of multiple chained generators and discriminators. The input of each stage of the chain is the output of the previous stage. One of the major differences between StackGAN-v2 and StackGAN-v1 is that these stages are trained jointly, whereas in StackGAN-v1, they are trained independently.

In our experiments, we found that the method by which we encode the textual descriptions can indeed have a big impact on the quality of the generated images. Here, we discuss the text embedding that we applied. 

\textbf{Text embedding:}\\
Both \textbf{StackGAN-v1} and \textbf{StackGAN-v2} condition the image generation process on  $\varphi_t$, i.e. the text embedding of the corresponding image description generated from a pre-trained char-CNN-RNN encoder~\citep{DBLP:journals/corr/ReedASL16}. It is important for the embedding of the description to correctly relate to the visual contents of the product image. We conducted our experiments using different encoders from a wide range of complexity, namely averaging word vectors, concatenating word vectors, a slightly modified encoder from the Transformer architecture~\cite{VaswaniSPUJGKP17} and a bidirectional LSTM~\cite{birnns}. 

We experimented with both pre-training these models \footnote{The pre-training step consisted of training the encoder to perform a classification task: given the item description predict its category.} and jointly training them with the GAN network. In the case of the Transformer's encoder and bi-LSTM, the text embedding $\varphi_t$ is the output of the encoder of the Transformer, and the projected concatenation of the last hidden state of the forward and backward LSTM respectively. The final text embedding size for the Transformer is $1500$ and $1024$ for bi-LSTM. 

\begin{figure*}
\centering
\includegraphics[height=0.45\textheight]{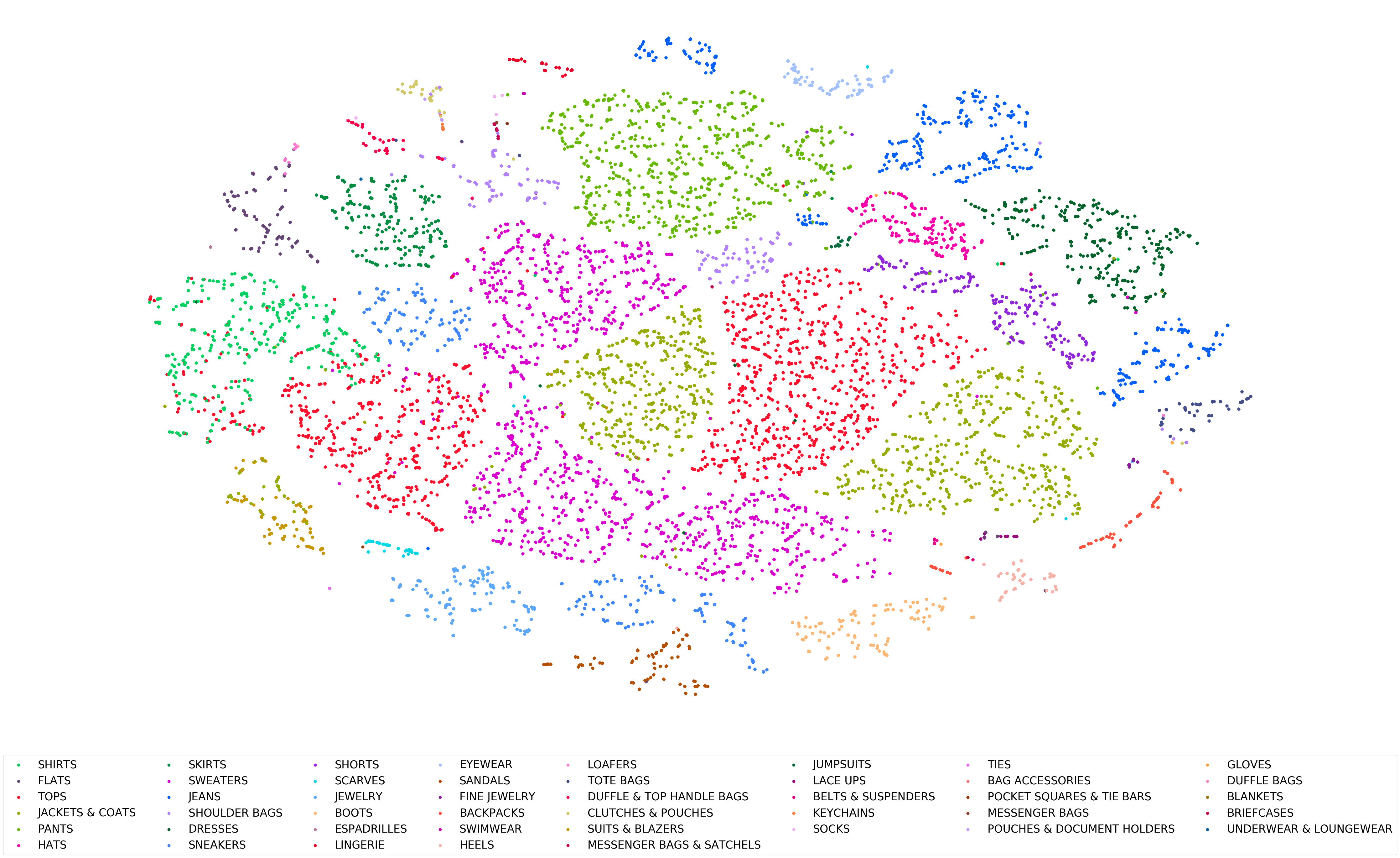}
\caption{t-SNE visualization of the validation text embedding obtained by the bi-LSTM encoder}
\label{valid_tsne}
\end{figure*}

We arrived at three conclusions based on our empirical experiments. First, we found that the bi-LSTM model achieves the highest category classification accuracy on the validation dataset in the pre-training process. As can be seen in Figure \ref{valid_tsne}, the t-SNE ~\cite{vanDerMaaten2008} visualization of text embeddings shows relatively good separation of the categories. Secondly, we found that irrespective of the encoder architecture, pre-training the encoder model results in better correspondence between the descriptions and generated images. Finally, we found that overall, using the pre-trained bi-LSTM with fixed weights as the encoder leads to better results both visually and quantitatively.

The Inception scores reported in table~\ref{inception} were obtained with the pre-trained bi-LSTM encoder (with fixed weights during the training of GAN).

\begin{figure}[ht]
    \centering
    \includegraphics[height=10cm]{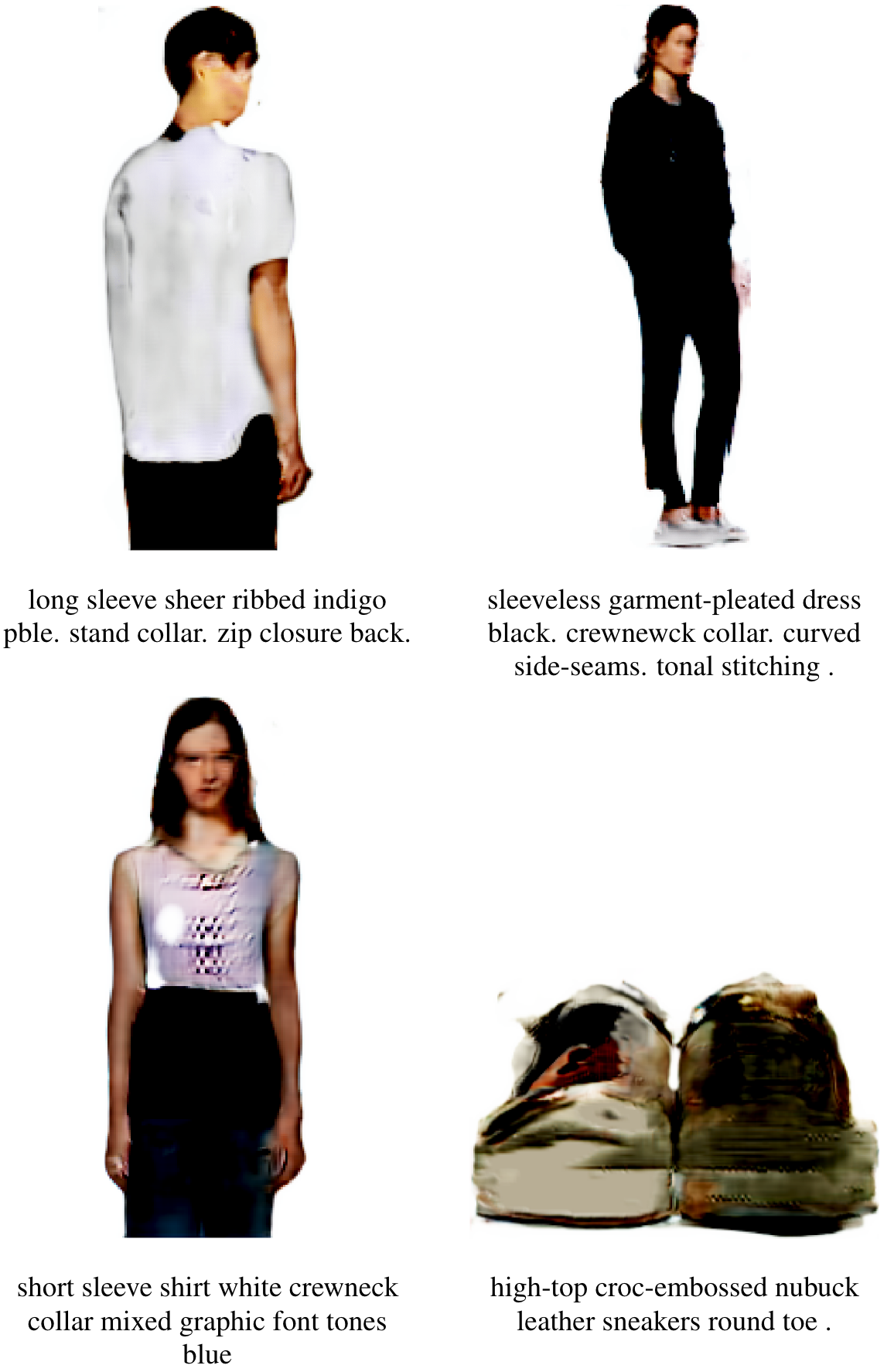}
    \caption{Images generated  from the \textbf{StackGAN-v1} model with pre-trained bi-LSTM text encoder.}
    \label{StackGAN-v1}
\end{figure}   

\begin{figure}[ht]
    \centering
    \includegraphics[height=10cm]{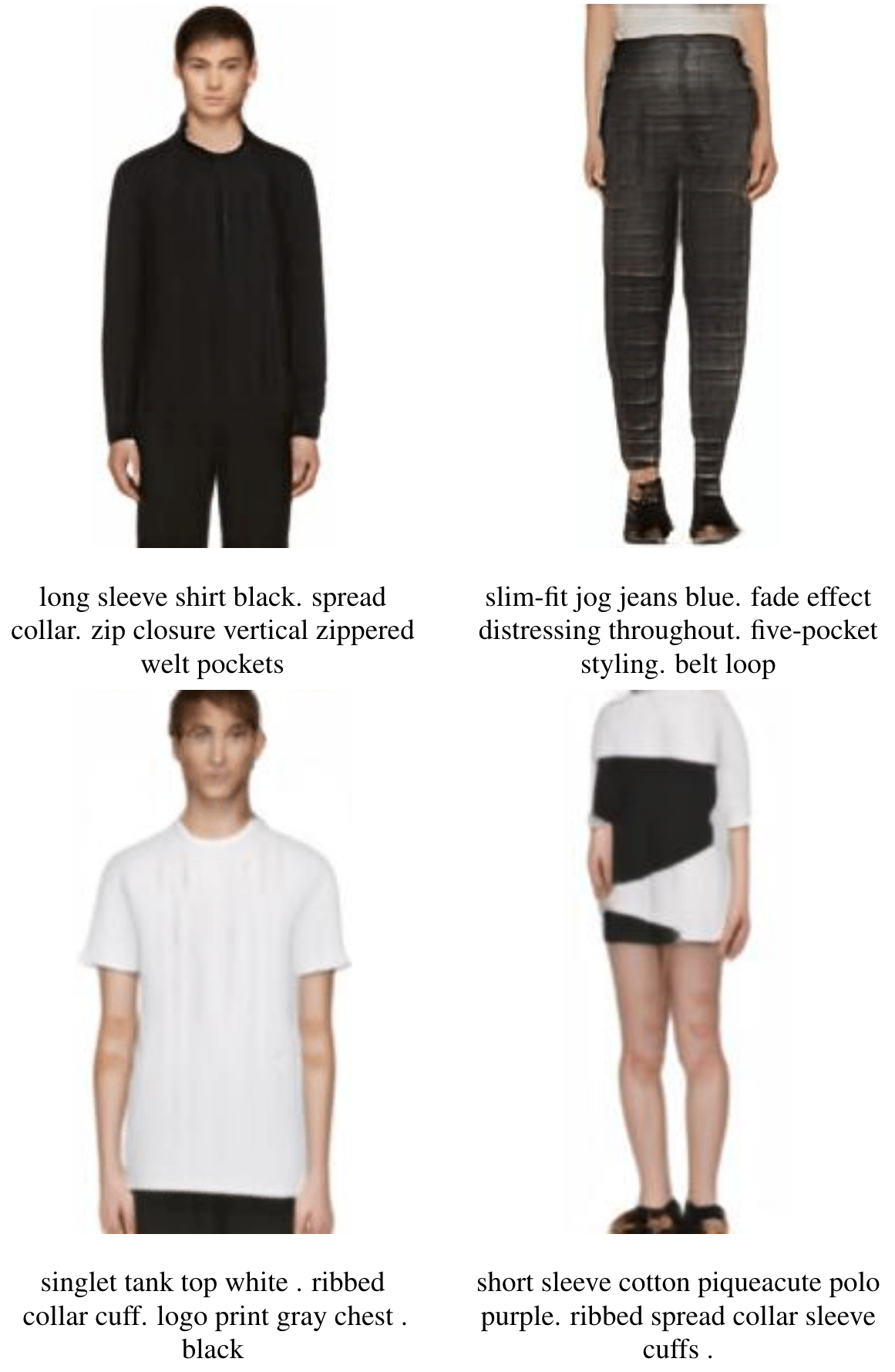}
    \caption{Images generated from the \textbf{StackGAN-v2} model with pre-trained bi-LSTM text encoder.}
    \label{StackGAN-v2}
\end{figure}   

\textbf{Implementation details:}\\
Throughout all the experiments, the descriptions were lowercased, tokenized and cleared of stop words\footnote{The python \href{http://www.nltk.org}{NLTK} module was used to tokenize the descriptions by word and remove stop words}. We used the first 15 tokens of the descriptions as the input sequence to the encoder model.  

\textbf{StackGAN-v1:}
We used the same overall architecture as~\cite{zhang2017stackgan} \footnote{We used the code provided by the authors of the StackGAN-v1 paper in github \href{https://github.com/hanzhanggit/StackGAN-Pytorch}{https://github.com/hanzhanggit/StackGAN-Pytorch}}. The first stage was trained for $80$ epochs, and the second stage was trained for $185$ epochs. The results can be seen in Figure~\ref{StackGAN-v1}.

%

\textbf{StackGAN-v2:}
After careful experimentation, we ended up using the same architecture and hyper-parameters as~\cite{DBLP:journals/corr/abs-1710-10916} \footnote{We used the code provided by the authors of the StackGAN-v2 paper: \href{https://github.com/hanzhanggit/StackGAN-v2}{https://github.com/hanzhanggit/StackGAN-v2}}. The results can be seen in Figure~\ref{StackGAN-v2}.

\begin{table}
\centering
\begin{tabular}{lc} 
\hline
                            & Inception Score  \\ 
\hline
Fashion Real data $256\times 256$   & $9.71 \pm 2.14$     \\
StackGAN-v1~\cite{zhang2017stackgan}                 & $6.50 \pm 0.05$     \\
StackGAN-v2~\cite{zhang2017stackgan}                 & $5.54 \pm 0.07$     \\
P-GAN~\cite{karras2017progressive} & $7.91 \pm 0.15$     \\
\hline
\end{tabular}
\caption{Inception Scores on the validation set, i.e: trained on the Fashion train set.}
\label{inception}
\end{table}

We can observe in the Table~\ref{inception} that first of all, the Inception Score of the StackGAN-V1 is better than StackGAN-V2, while the quality of the images in the StackGAN-V2 is better and the reason is due to a significant mode-collapse that we were faced to in StackGAN-V2. Another interesting point, is the fact that most of the faces in StackGAN-V1 and StackGan-V2 are blurry. It suggests that since the images are conditioned on the text, the model is focusing more on clothing material than face information.

\section{Conclusion}\label{conclusion}
Recent progress in generative modeling techniques has great potential to give designers tools for rapidly visualizing and modifying ideas. While recent advances in generative models 
can be used to generate images of unprecedented realism, the quality of images generated from textual descriptions has so-far remained far from realistic. We believe that the lack of good datasets for this task has made it difficult to develop models for this task. In this paper, we have introduced a new Fashion themed text-to-image generation dataset, with high-quality images and extensive annotations provided by fashion experts. We provided results for 2 experiments: generating high-resolution images without providing textual descriptions as input, and generating realistic images conditioned on product description using the Fashion dataset as training data. We provide experiments with StackGAN-v1 and StackGAN-v2 models using various text encoders.

To help stimulate further research on conditional generative models, we release our dataset as part of a challenge. Detailed submission instructions are provided and our API computes the inception score (trained on the Fashion dataset). Submissions with the highest quality images as judged by human evaluators will be selected as winners in the challenge organized around this new dataset.

\section{Acknowledgement}\label{acknowledgement}
We present our special thank to Alex Shee, for his help and support. We also thank Timnit Gebru and Archy de Berker, for assistance with comments that greatly improved the manuscript.
We would also like to show our gratitude to Chelsea Moran, Valerie Becaert, Vincent Hoe-Tin-Noe, Misha Benjamin, Pedro Oliveira Pinheiro, David Vazquez, Francis Duplessis, Ishmael Belghazi, Caroline Bourbonniere and Xavier Snelgrove for their support and feedback during the course of this research. 
We would also like to thank SSENSE for open sourcing their data to the research community.

\nocite{langley00}

\bibliography{example_paper}
\bibliographystyle{icml2018}

\appendix
\cleardoublepage
\newpage
\tiny{
  \tablefirsthead{%
    \hline
    \multicolumn{1}{l}{\scriptsize \textbf{Categories}} &
    \multicolumn{1}{l}{\scriptsize \textbf{Subcategories}} &
    \multicolumn{1}{r}{\scriptsize \textbf{Number of images}} \\
    \hline}
  \tablehead{%
    \hline
    \multicolumn{1}{l}{\scriptsize \textbf{Categories}} &
    \multicolumn{1}{l}{\scriptsize \textbf{Subcategories}} &
    \multicolumn{1}{r}{\scriptsize \textbf{Number of images}} \\
    \hline}
  \tabletail{\bottomrule}
  \bottomcaption{Number of images per category and subcategory in the training set.}
 
  \label{subcat}
  \begin{supertabular}{L{3cm}L{2cm}R{2cm}}
  BACKPACKS & BACKPACKS &               2858 \\
  BAG ACCESSORIES & BAG ACCESSORIES &                110 \\
  BELTS \& SUSPENDERS & BELTS \& SUSPENDERS &                335 \\
  BLANKETS & BLANKETS &                 17 \\
  BOAT SHOES \& MOCCASINS & BOAT SHOES \& MOCCASINS &                 12 \\
  BOOTS & ANKLE BOOTS &               1975 \\
                            & MID-CALF BOOTS &                737 \\
                            & CHELSEA BOOTS &                562 \\
                            & LACE-UP BOOTS &                441 \\
                            & TALL BOOTS &                407 \\
                            & ZIP UP \& BUCKLED BOOTS &                327 \\
                            & DESERT BOOTS &                 30 \\
                            & BIKER \& COMBAT BOOTS &                 24 \\
                            & WINGTIP BOOTS &                 12 \\
  BRIEFCASES & BRIEFCASES &                 71 \\
  CLUTCHES \& POUCHES & POUCHES &                640 \\
                            & CLUTCHES &                480 \\
  DRESSES & SHORT DRESSES &               4819 \\
                            & MID LENGTH DRESSES &               3809 \\
                            & LONG DRESSES &                830 \\
  DUFFLE \& TOP HANDLE BAGS & DUFFLE \& TOP HANDLE BAGS &               1533 \\
  DUFFLE BAGS & DUFFLE BAGS &                120 \\
  ESPADRILLES & ESPADRILLES &                119 \\
  EYEWEAR & SUNGLASSES &               1706 \\
                            & GLASSES &                788 \\
  FINE JEWELRY & RINGS &                215 \\
                            & EARRINGS &                200 \\
                            & BRACELETS &                 36 \\
                            & NECKLACES &                 24 \\
  FLATS & SLIPPERS \& LOAFERS &               1680 \\
                            & BALLERINA FLATS &                861 \\
                            & LACE UPS \& OXFORDS &                684 \\
                            & ESPADRILLES &                630 \\
  GLOVES & GLOVES &                 66 \\
  HATS & CAPS \& FLAT CAPS &               1846 \\
                            & BEANIES &               1219 \\
                            & STRUCTURED HATS &                423 \\
                            & FEDORAS \& PANAMA HATS &                367 \\
                            & CAPS &                360 \\
                            & HEADBANDS \& HAIR ACCESSORIES &                174 \\
                            & BEACH HATS &                116 \\
                            & AVIATOR &                 40 \\
  HEELS                     & HEELS &               1931 \\

 JACKETS \& COATS           & COATS &               7190 \\
                            & JACKETS &               6373 \\
                            & BOMBERS &               5864 \\
                            & DOWN &               5359 \\
                            & DENIM JACKETS &               3177 \\
                            & LEATHER JACKETS &               3080 \\
                            & FUR \& SHEARLING &               1161 \\
                            & BLAZERS &               1111 \\
                            & VESTS &                794 \\
                            & TRENCH COATS &                685 \\
                            & PEACOATS &                241 \\
  JEANS & JEANS &              13586 \\
  JEWELRY & BRACELETS &               1926 \\
                            & EARRINGS &               1173 \\
                            & RINGS &               1014 \\
                            & NECKLACES &                794 \\
                            & PINS &                 17 \\
                            & BROOCHES &                  9 \\
  JUMPSUITS & JUMPSUITS &                610 \\
  KEYCHAINS & KEYCHAINS &                202 \\
  LACE UPS & LACE UPS &                631 \\[5mm]
  LINGERIE & BRAS &                636 \\
                            & BRIEFS &                259 \\
                            & THONGS &                252 \\
                            & SLEEPWEAR &                 89 \\
                            & BOY SHORTS &                 56 \\
                            & TANKS &                 14 \\
                            & ROBES &                  8 \\
  LOAFERS & LOAFERS &                481 \\
  MESSENGER BAGS & MESSENGER BAGS &                 88 \\
  MESSENGER BAGS \& SATCHELS & MESSENGER BAGS \& SATCHELS &                530 \\
  MONKSTRAPS & MONKSTRAPS &                 29 \\
  PANTS & TROUSERS &              13590 \\
                            & SWEATPANTS &               5128 \\
                            & LOUNGE PANTS &               2565 \\
                            & CARGO PANTS &                677 \\
                            & LEGGINGS &                654 \\
                            & LEATHER PANTS &                616 \\
  POCKET SQUARES \& TIE BARS & POCKET SQUARES \& TIE BARS &                 23 \\
  POUCHES \& DOCUMENT HOLDERS & POUCHES \& DOCUMENT HOLDERS & 121 \\
  SANDALS & FLAT SANDALS &               1864 \\
                            & HEELED SANDALS &                937 \\
                            & SANDALS &                660 \\
                            & FLIP FLOPS &                 54 \\
  SCARVES & KNITS &                458 \\
                            & SCARVES &                432 \\
                            & SILKS \& CASHMERES &                183 \\
                            & FUR \& SHEARLING &                 65 \\
  SHIRTS & SHIRTS &              11398 \\
  SHORTS & SHORTS &               7416 \\
  SHOULDER BAGS & SHOULDER BAGS &               6952 \\
  SKIRTS & SHORT SKIRTS &               3262 \\
                            & MID LENGTH SKIRTS &               3182 \\
                            & LONG SKIRTS &                465 \\
  SNEAKERS & LOW TOP SNEAKERS &               7810 \\
                            & HIGH TOP SNEAKERS &               2620 \\
                            & WEDGE SNEAKERS &                 48 \\
  SOCKS & SOCKS &                250 \\
  SUITS \& BLAZERS & BLAZERS &               2560 \\
                            & SUITS &                171 \\
                            & TUXEDOS &                 73 \\
                            & WAISTCOATS &                 41 \\
  SWEATERS & CREWNECKS &              13399 \\
                            & SWEATSHIRTS &              12331 \\
                            & HOODIES \& ZIPUPS &              10388 \\
                            & TURTLENECKS &               3775 \\
                            & CARDIGANS &               3421 \\
                            & V-NECKS &               1261 \\
                            & SHAWLNECKS &                 16 \\
  SWIMWEAR & BIKINIS &                159 \\
                            & SWIMSUITS &                 21 \\
                            & ONE-PIECE &                 12 \\
                            & COVER UPS &                  9 \\
  TIES & NECK TIES &                240 \\
                            & BOW TIES &                 13 \\
  TOPS & T-SHIRTS &              33004 \\
                            & SHIRTS &               4563 \\
                            & BLOUSES &               3655 \\
                            & TANK TOPS \& CAMISOLES &               3052 \\
                            & POLOS &               2117 \\
                            & TANK TOPS &                642 \\
                            & BODYSUITS &                526 \\
                            & HENLEYS &                205 \\
  TOTE BAGS & TOTE BAGS &               2281 \\
  TRAVEL BAGS & TRAVEL BAGS &                 29 \\
  UNDERWEAR \& LOUNGEWEAR & ROBES &                 17 \\
                            & BOXERS &                  6 \\
\end{supertabular}	
}

\small
\end{document}